\documentclass{article}
\usepackage{spconf,amsmath,graphicx}

\usepackage{booktabs}       
\usepackage{multirow}
\usepackage{amsfonts}       
\usepackage{nicefrac}       
\usepackage{microtype}      
\usepackage{subfigure}
\usepackage{graphicx}
\usepackage{mathrsfs}
\usepackage{helvet}
\usepackage{courier}
\usepackage{amsmath}
\usepackage{amssymb}
\usepackage{enumitem}
\usepackage{setspace}

\usepackage{fancyhdr}       

\title{Selective Convolutional Network: An Efficient \\Object Detector with Ignoring Background}
\name{Hefei Ling$^{\star \dagger}$, Yangyang Qin$^{\star}$\thanks{$^{\star}$Equal contribution $^{\dagger}$Corresponding author}, Li Zhang, Yuxuan Shi, Ping Li\thanks{
This work was supported in part by the Natural Science Foundation of China under Grant U1536203 and 61972169, in part by the National key research and development program of China(2016QY01W0200),  in part by the Major Scientific and Technological Project of Hubei Province (2018AAA068 and 2019AAA051).
}}
\address{Department of Computer Science and Technology, Huazhong University of Science and Technology, CN}

\begin{document}
%
\maketitle
\thispagestyle{fancy}
\fancyhead{}
\lhead{}
\lfoot{\small ~\copyright~2020 IEEE. Published in the IEEE 2020 International Conference on Acoustics, Speech, and Signal Processing (ICASSP 2020), scheduled for 4-9 May, 2020, in Barcelona, Spain.}
\cfoot{}
\rfoot{}
\begin{abstract}
 It is well known that attention mechanisms can effectively improve the performance of many CNNs including object detectors. Instead of refining feature maps prevalently, we reduce the prohibitive computational complexity by a novel attempt at attention.  Therefore, we introduce an efficient object detector called \emph{Selective Convolutional Network} (SCN), which selectively calculates only on the locations that contain meaningful and conducive information. The basic idea is to exclude the insignificant background areas, which effectively reduces the computational cost especially during the feature extraction. To solve it, we design an elaborate structure with negligible overheads to guide the network where to look next. It's end-to-end trainable and easy-embedding. Without additional segmentation datasets, we explores two different train strategies including direct supervision and indirect supervision. Extensive experiments assess the performance on PASCAL VOC2007 and MS COCO detection datasets. Results show that SSD and Pelee integrated with our method averagely reduce the calculations in a range of 1/5 and 1/3 with slight loss of accuracy, demonstrating the feasibility of SCN. 
\end{abstract}


\section{INTRODUCTION}
\label{section_introduction}
With the development of deep learning, CNN-based detectors have occupied the dominant position in object detection. We prefer one-stage to two-stage detectors as they are fast and efficient in deploying on the ordinary computer even mobile devices, such as SSD \cite{liu2016ssd}. Even though, developers are often troubled by the expense of massive computational budget that results from the deep and wide architecture. To solve this challenging task, some innovative lightweight CNN models have been proposed in recent years, such as MobileNet \cite{sandler2018mobilenetv2}, Pelee \cite{wang2018pelee}, which can run inference on ordinary device in real time. What's more, there are many methods to compress
models, similar to our purpose, such as structured pruning \cite{he2017channel,yamamoto2018pcas}, knowledge distillation \cite{hinton2015distilling,zhu2019mask}, adaptive inference\cite{veit2018convolutional,wang2018skipnet}.

\begin{figure}
    \centering
    \subfigure[A bicycle on the street]{
        \begin{minipage}[t]{0.45\linewidth}
            \centering
            \includegraphics[height=0.8in]{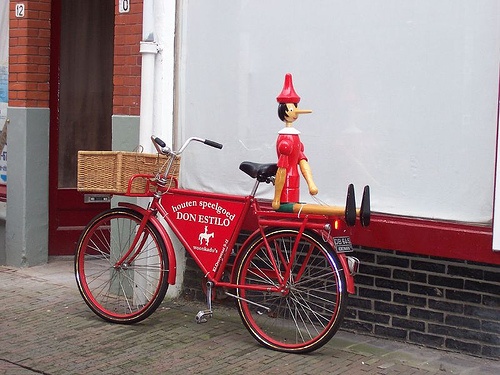}
        \end{minipage}%
    }
    ~
    \subfigure[Erase the surrounding]{
        \begin{minipage}[t]{0.45\linewidth}
            \centering
            \includegraphics[height=0.8in]{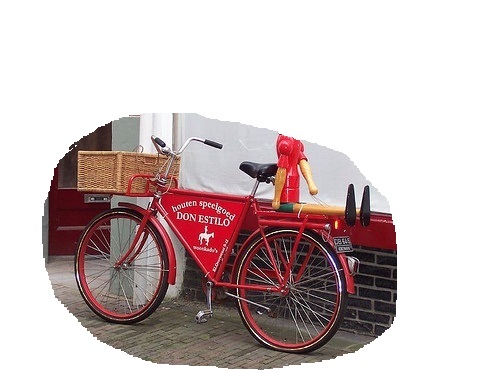}
        \end{minipage}
        \label{erase_bycle}
    }
    \caption{Erasing the surrounding information can't confuse us to seek out bicycles and should not hurt the performance of locating bicycles for detectors if their design is reasonable.}
    \label{bycle}
\vspace{-0.2cm}
\end{figure}

Existing object detection models always perform convolution calculation on the whole spatial information, which results in a large amount of computation. An important thing that has been overlooked in the past researches is that people will not look closely at all the goals in front of them, due to the efficient strategy that the human visual system has learned imperceptibly through the superior colliculus (SC) structure \cite{guo2014fast,white2017superior}.
Taking the street scene in Fig.~\ref{erase_bycle} as an example, the process of searching for bicycles will only speed up under the influence of SC structure, rather than lose targets just because of the missing surrounding information.
Guided by this illuminating observation, it is possible to detect specified objects only relying on the partial but significant information mostly from the foreground.
Though DCN \cite{dai2017deformable} uses the deformable convolution with spatial domain offsets to focus on the specific objects instead of the adjacent background, dispensable activations on the background locations will be also calculated. And many works \cite{woo2018cbam,wang2018non,wang2017face} only use attention mechanisms to enhance certain features, which violates the original intention to decrease the size of search spaces.

Our motivation is to avoid the generation of spatial information redundancy and it resembles a sophisticated spatial pruning more in line with the efficient human visual system to some extent.
We make two major contributions as follows:
\begin{itemize}
\item We propose a new efficient method called SCN to selectively perform convolution according to the generated foreground mask, which is specially tailored to object detection.
     Our method has the following merits: i) Easy to embed. ii) Almost no accuracy loss.
\item We explore some effortless strategies to get the foreground mask, including a hand-crafted extended branch called Selective Module, and two entirely different strategies to train the module without requiring use of segmentation datasets, as described in Section \ref{section_supervision}.
\end{itemize}



\section{SELECTIVE CONVOLUTIONAL NETWORK}
\subsection{Overview}
\begin{figure}[htbp]
    \centering
    \includegraphics[width=1\linewidth]{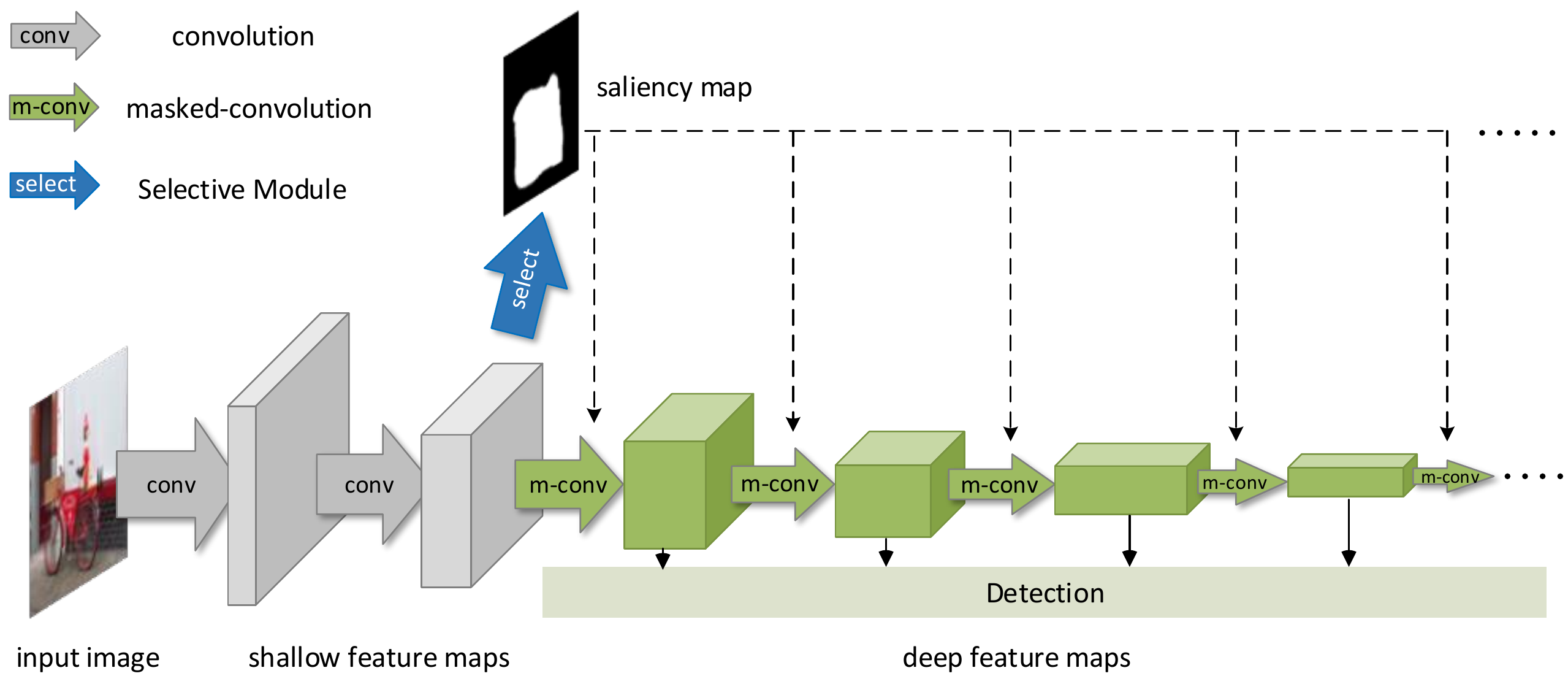}
    \caption{An overview of our SCN. The added saliency map is inferred from the former shallow features, and then instruct the subsequent layers to efficiently extract the specified features.}
    \label{network_overview}
\end{figure}
\begin{figure}[tbp]
    \centering
    \includegraphics[width=0.8\linewidth]{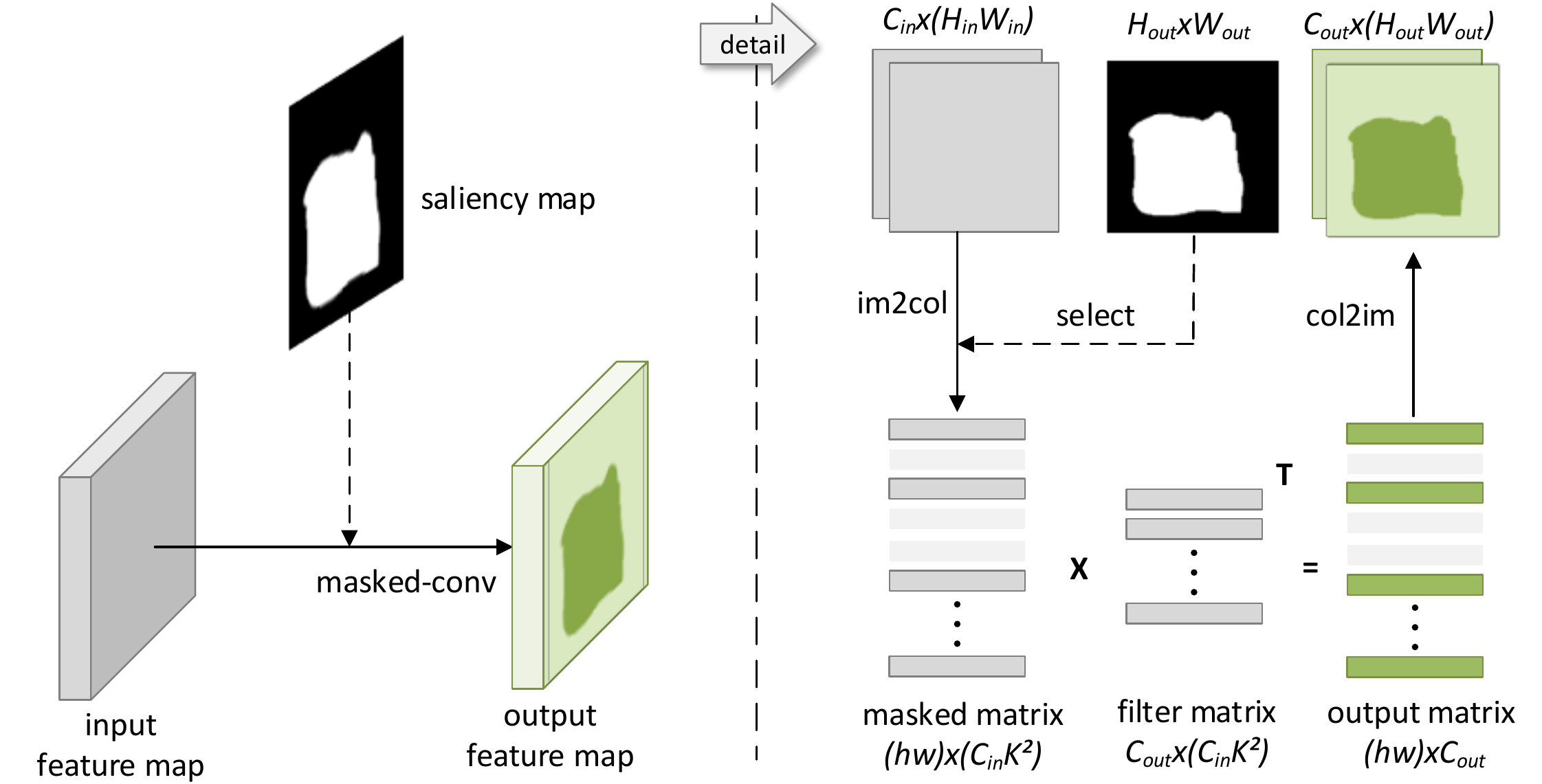}
    \caption{After $im2col$, the elements in $H_{in}\times W_{in}$ rows of the feature matrix represent the features at corresponding spatial location, and then are selected into $h\times w$ ones to join calculation by saliency map, as $h\times w$ is the number of non-zero entries in saliency map. Finally, uncalculated locations (light olive) in output matrix will be filled into 0 for restoring shape.}
    \label{mask_conv}
\end{figure}

SCN is designed to bring down the cost of computation on dispensable spatial locations. Applied to the popular SSD \cite{liu2016ssd}, an overview of our network structure can be found in Fig.~\ref{network_overview}. Few changes have been made to the original framework except an extended branch called Selective Module for predicting saliency maps. Saliency is to filter the visual information and select interesting ones for further processing \cite{guo2014fast}. In our overall architecture, saliency maps are rapidly generated from the former shallow feature maps and guide the latter layers where to do calculation through masked-convolutions \cite{song2018beyond}, where we extend saliency as a binary location-guided mask. As an extended branch, the Selective Module is a tiny architecture and shares features with the shallower trunk branch, inspired by Mask R-CNN \cite{he2017mask}. We cautiously design this branch so that it won't put the brakes on the execution of the trunk branch. Module details are described in Section \ref{section_SM}. After capturing the saliency map, subsequent layers adopt the masked-convolution instead of the vanilla convolution to reduce computation cost extremely effectively as outlined in Fig.~\ref{mask_conv}, which would lose little information on key locations. Moreover, ignored spatial locations will not predict the detection results for decreasing false positives. For a clearer explanation, the process of masked-convolution can be formulated as:
\begin{eqnarray}
x_{input}^{'} & = & \mathcal{F}(im2col(x_{input}), m_{s})\\
x_{output}^{'} & = & wx_{input}^{'} \\
x_{output} & = & col2im(\mathcal{G}(x_{output}^{'}, m_{s}))
\end{eqnarray}
where $x_{input}$ and $x_{output}$ are input and output respectively, $w$ stands for the filter matrix, and $m_{s}$ is the corresponding saliency map.
$im2col$ \cite{im2col} converts the feature maps into matrix and $col2im$ is the opposite.
$\mathcal{F}(\cdot)$ and $\mathcal{G}(\cdot)$ indicate the selective and the scatter function, respectively.
What's more, different size masks need to be generated corresponding to different sizes of feature maps, and our experiments show that using downsampling with stride-convolution from the original size mask is better than simple pooling to perfect adaption because of the training strategy seen in Section \ref{section_supervision}.

\subsection{Selective module}
\label{section_SM}

\begin{figure}[htbp]
    \centering
    \includegraphics[width=0.6\linewidth]{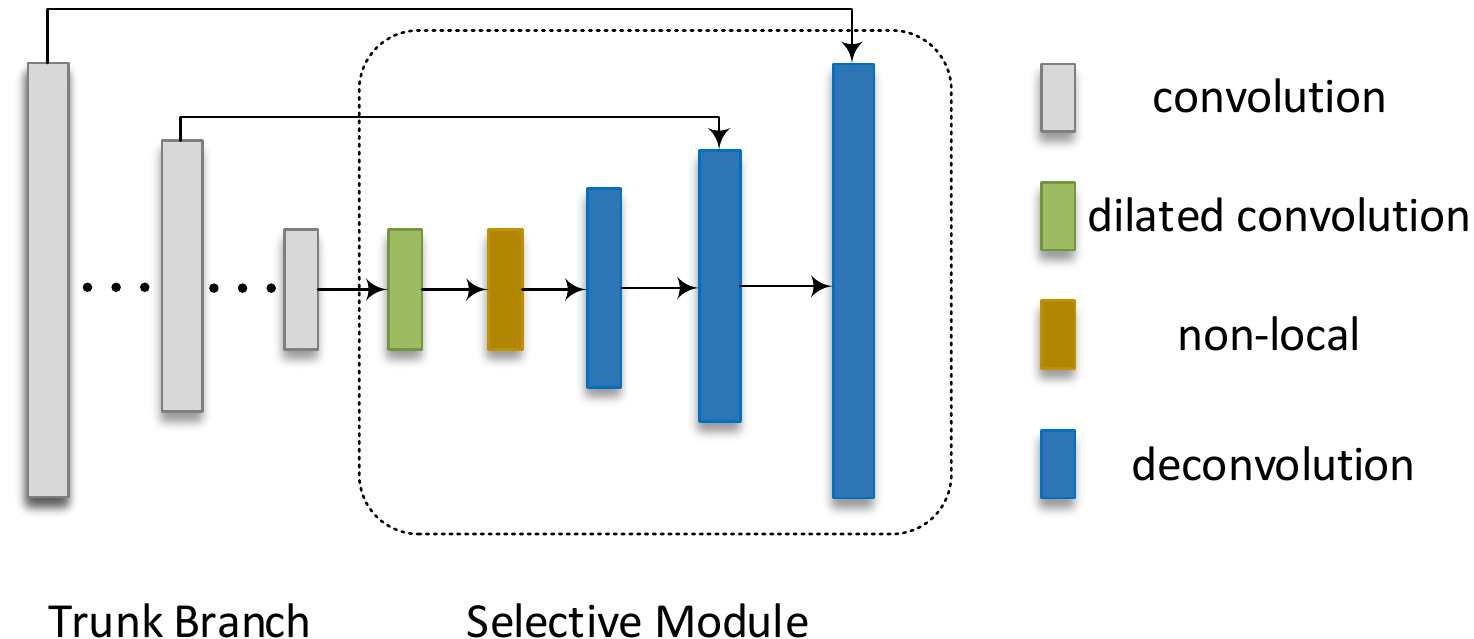}
    \caption{Illustration of Selective Module.}
    \label{selective_module}
\end{figure}

Selective Module is an essential component of SCN for guiding the network to do lightweight computing. Although it is not a simple matter to get the saliency map, we argue that our elaborate structure with less computational cost can achieve satisfactory results, as illustrated in Fig.~\ref{selective_module}.
Motivated by \cite{long2015fully,badrinarayanan2017segnet,noh2015learning}, we adopt an encoder-decoder structure as the transformer for pixel-wise segmentation and use deconvolution as the manner of decoder upsampling, attaching skip architecture. Instead of using preprocessing subnetwork, we propose a plug-and-play network branch to generate foreground masks.
The new branch is attached to the shallower trunk branch and shares the bottom-up structure with the trunk to exceedingly reduce the additionally introduced computation, which can almost eliminate the encoder part. However, the semantics of shared feature is insufficient, and the receptive field is not big enough because of fewer strides. To address this problem, we adopt dilated convolution \cite{yu2015multi} and non-local \cite{wang2018non} to expand the receptive field and aggregation context information for better semantics. The number of channels in the Selective Module is quite small compared with the trunk and the convolutions can be replaced by the depthwise separable unit.
%

Selective Module will generate saliency maps from given feature maps, and which feature to be chosen as input is critical. As discussed above, the selected location determines the capability of the encoder network. Embedded too shallow will make it hard to get desired saliency maps. Meanwhile, too deep location hardly reduces computation. Regard to the VGG-16, a backbone with $300\times300$ pixels resolution, we use the $75\times75$ and the $38\times38$ size output feature maps from the Pool2 and the Pool3 layer for embedded location comparison.

\subsection{Direct supervision or indirect supervision}
\label{section_supervision}

{\bf Direct Supervision.}
An obvious way to train our proposed Selective Module is to supervise the mask results directly.
Of course, we won't use any segmentation datasets for the sake of fairness. All the ground-truth masks are generated from the bounding boxes. Specifically, we regard the areas inside the bounding boxes as foreground mask 1 and the others as background mask 0. Therefore every ground-truth foreground mask is a square block. The mask value denotes the need for further inference. Note that we experimentally expand one stride size around the ground-truth foreground masks, which can mitigate the harm from the deviation of predicted masks and retain some contextual information.
In terms of the loss function,
in addition to the common classification loss $L_{c}$ and localization loss $L_{l}$ for detection, we regard saliency as foreground segmentation and define a dedicated per-pixel sigmoid cross entropy as saliency mask loss $L_{m}$ for pixel-wise classification to achieve our goals.
Then use a threshold function $(\psi=0.5)$ for converting the probability map to the binary saliency we need.
And these three loss functions compose of a multi-task loss function as $L = L_{c} + {\lambda}_{1} L_{l} + {\lambda}_{2} L_{m}$ to jointly optimize parameters. $L_{c}$ and $L_{l}$ are identical as the softmax loss and the smooth L1 loss defined in SSD \cite{liu2016ssd}, and the saliency mask loss $L_{m}$ is defined as:
$L_{m} = \frac{1}{N}\sum_{k=1}^{N}L_{ce}(m_{k} , m_{k}^{*}) $
where $N$ denotes the total number of the coordinates in all saliency masks. $L_{ce}$ is the binary cross-entropy loss. $m_{k}$ is the activation of each coordinate in saliency map, and $m_{k}^{*}$ is the corresponding ground-truth mask described above. Like other multi-task loss, ${\lambda}_{1}$ and ${\lambda}_{2}$ commonly aim to balance the three terms. Here we simply set ${\lambda}_{1} = {\lambda}_{2} = 1$. Different from the general way, ignored locations won't calculate classification and regression loss to focus the training target on useful foreground locations. 

{\bf Indirect Supervision.}
We can also train the Selective Module in an unsupervised fashion \cite{woo2018cbam,wang2018non}.
Predicting detection results on each spatial location of specified feature maps is a key characteristic of one-stage detectors \cite{liu2016ssd,dai2017deformable}, and different gradients flow at different locations during training. Therefore, it's possible to train and make the mask generation most beneficial to the prediction results without explicit supervision. The supervision signal of the Selective Module comes entirely from the final classification and detection loss. 
We think this training strategy leads to better accuracy. To prevent the gradient explosion of the crowded connection to the backbone when training, only the gradient in the guided layers close to detection heads will flow back. And multiplying mask with output feature maps instead of masked-convolution is more conducive to gradient flow during the training phase.

\section{EXPERIMENTS}
\label{section_experiments}

\subsection{Implementation details}
We implement all models on the PyTorch framework with a uniform input resolution of $300\times300$. For a fair comparison, the experiment settings and training strategies are all similar to the original SSD \cite{liu2016ssd} and no additional tricks except adding Batch Normalization for the training convenience. We use the SGD solver with a weight decay of 0.0005 and a momentum of 0.9. First, the models adopt the warm-up strategy for the first 5 epochs. Subsequently, we set the learning rate to 0.01 initially and then use a step decay strategy. The batch size is set to 32. Comparing with the original model, we double the number of iterations to guarantee sufficient training for the introduction of the additional saliency task.

\subsection{Results on PASCAL VOC}
PASCAL VOC dataset consists of natural images with 20 classes. We train all models on the union set of VOC 2007 trainval and VOC 2012 trainval, and test on VOC 2007 test set.

\subsubsection{Design choice}
\label{section_design_choice}
Table~\ref{design-choices} shows the performance of various embedded locations and supervision strategies, including reduced computational effort, measured in \emph{floating point operations} (FLOPs), and accuracy indicator. Since different images have different proportions of background, we take the average of lightweight indicators.
We can see that indirect supervision leads to higher accuracy, but they tend to be more conservative and don't dare to ignore too much complex background. 
While the direct supervision strategy is exactly the opposite. It encourages models to boldly ignore the background under the guidance. From another perspective, $38\times38$ location embedding shows exciting results with just 0.1\% to 0.2\% degradation. While deeper embedded location can reduce more calculations, observed from the comparison of the embedded location.

\begin{table}[htbp]
\setlength{\tabcolsep}{1.3mm}{
  \caption{Evaluation results of various \textbf{design choices} on VOC. *: The baseline is the pre-trained SSD under ImageNet.}
  \label{design-choices}
  \centering
  \begin{tabular}{ccccc}
    \toprule
    \multicolumn{2}{c}{Design}  &\multirow{2}*{GFLOPs}  & \multirow{2}*{Reduced}  & \multirow{2}*{mAP (\%)}\\
    \cmidrule(r){1-2}
    Location   & Supervision    & & & \\
    \midrule
    \hline
    \multicolumn{2}{c}{Baseline*}      &31.78 &- &79.1 \\
    \hline
    \multirow{3}*{$75\times75$}   & Indirect        &25.09  &6.69(21.0\%) &77.6           \\
                                    & Direct        &21.84  &{\bf9.94(31.3\%)} &77.4     \\
    \hline
    \multirow{2}*{$38\times38$}   & Indirect        &28.55  &3.23(10.2\%) &{\bf79.0}              \\
                                    & Direct        &25.71  &6.07(19.1\%)  &78.9              \\
    \bottomrule
  \end{tabular}
}
\end{table}

\begin{figure}[htbp]
    \centering
    \includegraphics[width=1\linewidth]{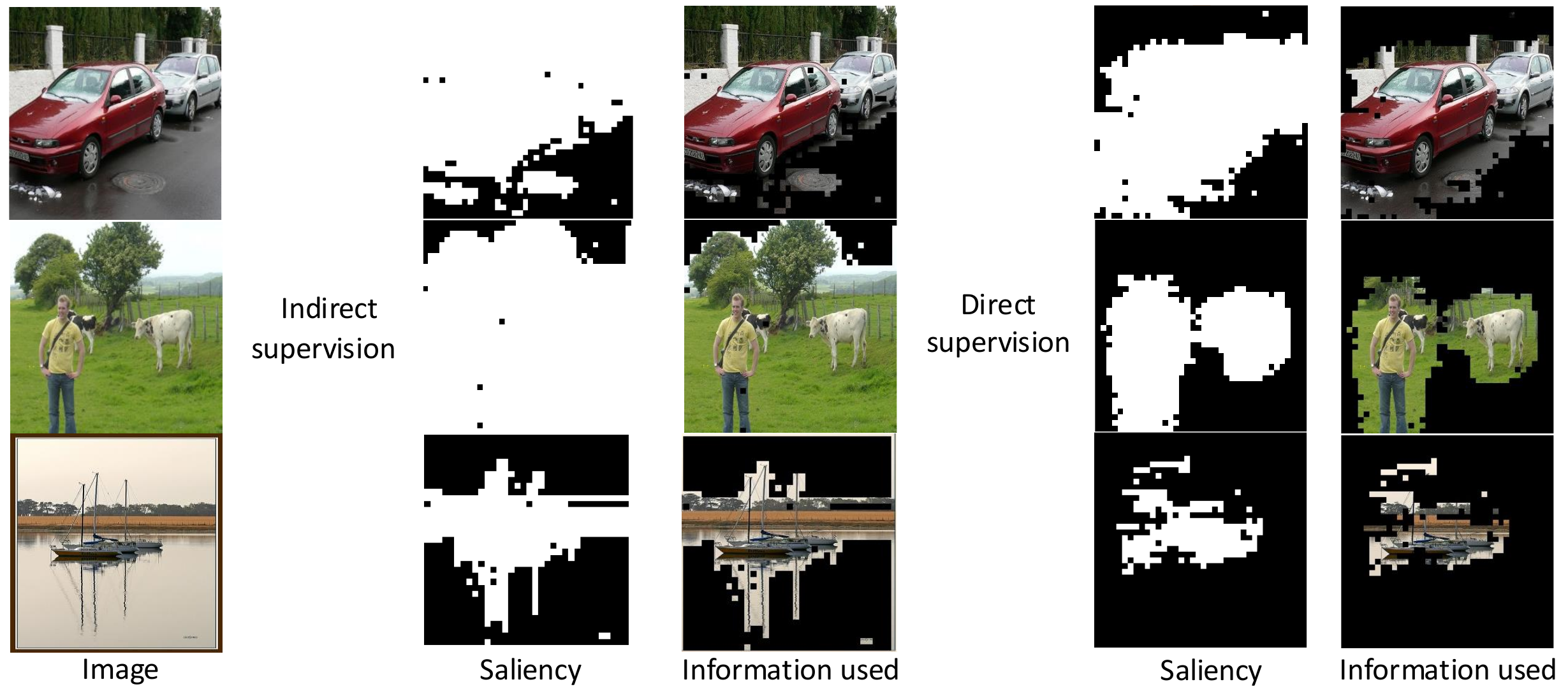}
    \caption{Visualization results during processing. For each row, we show an input image, two sets of saliency maps and corresponding information scope, where the maps corresponding from indirect supervision and direct supervision, respectively.}
    \label{examples}
\end{figure}

For more intuitive, some illustrative examples of inferred saliency map can be found in Fig.~\ref{examples}. The upper two rows show the conservatism of the indirect supervision strategy. Without guidance like direct supervision, the network hardly ignores the complicated background, because small objects are possible to hide in them, such as a cow hidden behind the man in the second row of Fig.~\ref{examples}.

\subsubsection{Ablation study}
\label{section_ablation}
As shown in Table~\ref{ablation-study}, we use $38\times38$ location embedding models as a baseline for ablation study.
 Results show that models with indirect supervision are more sensitive to the ablation of complicated subcomponents.
 But contrary to our intuition, they don't pay off under direct supervision.
 We consider it’s a confusing effect caused by the redundant and coarse mask supervision generated by the bounding box. As a matter of fact, the coarse ground-truth masks covered a lot of wrong locations, which make the network hesitate at the edge of objects. In other words, there is a contradiction between good structural performance and unreliable supervision so that we can't design an overly complicated module.

\begin{table}
\setlength{\tabcolsep}{0.9mm}{
  \caption{\textbf{Ablation study} of Selective Module on VOC.}
  \label{ablation-study}
  \centering
  \begin{tabular}{ccccccc}

    \toprule
    &Supervision &\multicolumn{5}{c}{Selective Module of SCN}    \\

    \midrule
    \hline
    Deconv upsample &      &               &\checkmark    &\checkmark    &\checkmark    &\checkmark    \\
    Skip connection &       &               &               &\checkmark    &\checkmark    &\checkmark    \\
    Non-local       &       &\checkmark     &               &              &\checkmark    &\checkmark     \\
    Dilated-conv    &       &\checkmark     &               &               &               &\checkmark   \\
    \hline
    \multirow{2}*{mAP(\%)} &Indirect&61.2&73.1&76.5&78.9&{\bf79.0} \\
                                                &Direct&77.5&78.9&{\bf78.9}&77.2&78.6\\
   \bottomrule
  \end{tabular}
  }
\end{table}

\begin{figure}[htbp]
    \centering
    \includegraphics[width=0.75\linewidth]{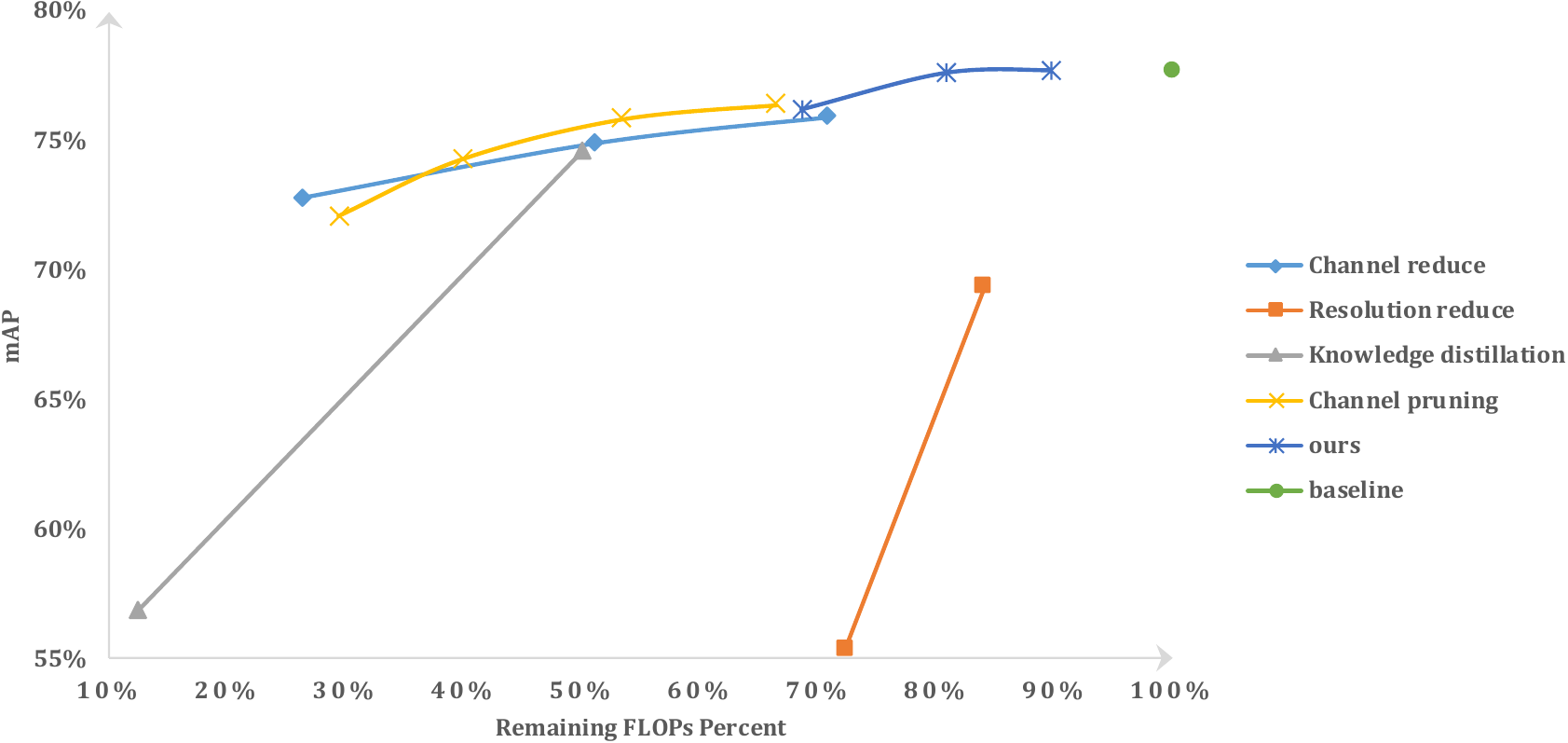}
    \caption{Comparison with other compression strategies on SSD.}
    \label{comparison}
\end{figure}
\subsubsection{Performance comparison}
\label{section_comparison}
The comparative methods are as follows: i) Channel reduce. ii) Resolution reduce. iii) Knowledge distillation with mask guided \cite{zhu2019mask}. iv) Channel pruning with LASSO-based channel selection\cite{he2017channel,wu2018pocketflow}.
Fig.~\ref{comparison} shows our accuracy degradation is very slight though the compression degree is general, which demonstrates the potential of our method.
All these strategies are not contradictory and can complement each other in practice.

\subsection{More challenging experiments}
In addition to the experiments of SSD on VOC, We further test on Pelee \cite{wang2018pelee}, a SOTA lightweight detector, and on more challenging MS COCO dataset, which contains more images and smaller objects. 
For the sake of simplicity, we only show the results of direct supervision in Table~\ref{COCO} and the indirect supervision situation is similar. The results are not much worse than SSD on VOC, which demonstrates the robustness of our method to handle more complicated examples.
\begin{table}
\setlength{\tabcolsep}{0.4mm}{
  \caption{Results of SCN on COCO dataset and Pelee detector.}
  \label{COCO}
  \centering
  \begin{tabular}{ccccccc}
    \toprule

    \multirow{2}*{Method} & \multirow{2}*{Data}& \multirow{2}*{GFLOPs}&\multirow{2}*{Reduced} &\multicolumn{3}{c}{AP (\%), IOU} \\
    \cmidrule{5-7}
    &  &  &  &0.5:0.95 &0.5 &0.75\\
    \midrule
    \hline
    Pelee &VOC     &1.18       &-   &-      &71.3 &-\\
    Pelee+SCN &VOC &0.79  &\textbf{0.39(33.1\%)}  &-&70.1&- \\
    \hline
    SSD  &COCO&35.58    &-  &26.9  &45.3  &28.1 \\
    SSD+SCN &COCO&27.45 &\textbf{8.13(22.9\%)}  &26.1  &44.3  &27.0    \\
    Pelee&COCO&1.25  &-   &22.6 &38.7 &23.1   \\
    Pelee+SCN&COCO&0.85 &\textbf{0.40(31.9\%)} &20.7 &36.2 &20.8   \\
    \bottomrule
  \end{tabular}
}
\end{table}

\section{CONCLUSION}

In this paper, we propose an efficient object detection method to eliminate redundant information processing on useless background locations. Experimental results show that our SCN can reduce the computational cost of SSD and Pelee in a range of 1/5 and 1/3 with little accuracy degradation within 2\%, including reducing the computational cost of SSD by about 20\% only with 0.2\% degradation.
 All these validate the feasibility of our novel method, and it can be easily integrated in not only ordinary detectors but also lightweight detectors.

\vfill\pagebreak

{
\bibliographystyle{IEEEbib}
\bibliography{mybib}

\begin{thebibliography}{10}

\bibitem{liu2016ssd}
Wei Liu, Dragomir Anguelov, Dumitru Erhan, Christian Szegedy, Scott Reed,
  Cheng-Yang Fu, and Alexander~C Berg,
\newblock ``Ssd: Single shot multibox detector,''
\newblock in {\em ECCV}. Springer, 2016, pp. 21--37.

\bibitem{sandler2018mobilenetv2}
Mark Sandler, Andrew Howard, Menglong Zhu, Andrey Zhmoginov, and Liang-Chieh
  Chen,
\newblock ``Mobilenetv2: Inverted residuals and linear bottlenecks,''
\newblock in {\em Proceedings of the IEEE Conference on Computer Vision and
  Pattern Recognition}, 2018, pp. 4510--4520.

\bibitem{wang2018pelee}
Robert~J Wang, Xiang Li, and Charles~X Ling,
\newblock ``Pelee: A real-time object detection system on mobile devices,''
\newblock in {\em Advances in Neural Information Processing Systems}, 2018, pp.
  1963--1972.

\bibitem{he2017channel}
Yihui He, Xiangyu Zhang, and Jian Sun,
\newblock ``Channel pruning for accelerating very deep neural networks,''
\newblock in {\em Proceedings of the IEEE ICCV}, 2017, pp. 1389--1397.

\bibitem{yamamoto2018pcas}
Kohei Yamamoto and Kurato Maeno,
\newblock ``Pcas: Pruning channels with attention statistics for deep network
  compression,''
\newblock {\em arXiv preprint arXiv:1806.05382}, 2018.

\bibitem{hinton2015distilling}
Geoffrey Hinton, Oriol Vinyals, and Jeff Dean,
\newblock ``Distilling the knowledge in a neural network,''
\newblock {\em arXiv preprint arXiv:1503.02531}, 2015.

\bibitem{zhu2019mask}
Yousong Zhu, Chaoyang Zhao, Chenxia Han, Jinqiao Wang, and Hanqing Lu,
\newblock ``Mask guided knowledge distillation for single shot detector,''
\newblock in {\em 2019 IEEE International Conference on Multimedia and Expo
  (ICME)}. IEEE, 2019, pp. 1732--1737.

\bibitem{veit2018convolutional}
Andreas Veit and Serge Belongie,
\newblock ``Convolutional networks with adaptive inference graphs,''
\newblock in {\em Proceedings of ECCV}, 2018, pp. 3--18.

\bibitem{wang2018skipnet}
Xin Wang, Fisher Yu, Zi-Yi Dou, Trevor Darrell, and Joseph~E Gonzalez,
\newblock ``Skipnet: Learning dynamic routing in convolutional networks,''
\newblock in {\em Proceedings of ECCV}, 2018, pp. 409--424.

\bibitem{guo2014fast}
Mingwei Guo, Yuzhou Zhao, Chenbin Zhang, and Zonghai Chen,
\newblock ``Fast object detection based on selective visual attention,''
\newblock {\em Neurocomputing}, vol. 144, pp. 184--197, 2014.

\bibitem{white2017superior}
Brian~J White, David~J Berg, Janis~Y Kan, Robert~A Marino, Laurent Itti, and
  Douglas~P Munoz,
\newblock ``Superior colliculus neurons encode a visual saliency map during
  free viewing of natural dynamic video,''
\newblock {\em Nature Communications}, vol. 8, pp. 14263, 2017.

\bibitem{dai2017deformable}
Jifeng Dai, Haozhi Qi, Yuwen Xiong, Yi~Li, Guodong Zhang, Han Hu, and Yichen
  Wei,
\newblock ``Deformable convolutional networks,''
\newblock in {\em Proceedings of the IEEE ICCV}, 2017, pp. 764--773.

\bibitem{woo2018cbam}
Sanghyun Woo, Jongchan Park, Joon-Young Lee, and In~So~Kweon,
\newblock ``Cbam: Convolutional block attention module,''
\newblock in {\em Proceedings of ECCV}, 2018, pp. 3--19.

\bibitem{wang2018non}
Xiaolong Wang, Ross Girshick, Abhinav Gupta, and Kaiming He,
\newblock ``Non-local neural networks,''
\newblock in {\em Proceedings of the IEEE Conference on Computer Vision and
  Pattern Recognition}, 2018, pp. 7794--7803.

\bibitem{wang2017face}
Jianfeng Wang, Ye~Yuan, and Gang Yu,
\newblock ``Face attention network: an effective face detector for the occluded
  faces,''
\newblock {\em arXiv preprint arXiv:1711.07246}, 2017.

\bibitem{song2018beyond}
Guanglu Song, Yu~Liu, Ming Jiang, Yujie Wang, Junjie Yan, and Biao Leng,
\newblock ``Beyond trade-off: Accelerate fcn-based face detector with higher
  accuracy,''
\newblock in {\em Proceedings of the IEEE Conference on Computer Vision and
  Pattern Recognition}, 2018, pp. 7756--7764.

\bibitem{he2017mask}
Kaiming He, Georgia Gkioxari, Piotr Doll{\'a}r, and Ross Girshick,
\newblock ``Mask r-cnn,''
\newblock in {\em Proceedings of the IEEE ICCV}, 2017, pp. 2961--2969.

\bibitem{im2col}
Kumar Chellapilla, Sidd Puri, and Patrice Simard,
\newblock ``High performance convolutional neural networks for document
  processing,''
\newblock 10 2006.

\bibitem{long2015fully}
Jonathan Long, Evan Shelhamer, and Trevor Darrell,
\newblock ``Fully convolutional networks for semantic segmentation,''
\newblock in {\em Proceedings of the IEEE conference on computer vision and
  pattern recognition}, 2015, pp. 3431--3440.

\bibitem{badrinarayanan2017segnet}
Vijay Badrinarayanan, Alex Kendall, and Roberto Cipolla,
\newblock ``Segnet: A deep convolutional encoder-decoder architecture for image
  segmentation,''
\newblock {\em IEEE transactions on pattern analysis and machine intelligence},
  vol. 39, no. 12, pp. 2481--2495, 2017.

\bibitem{noh2015learning}
Hyeonwoo Noh, Seunghoon Hong, and Bohyung Han,
\newblock ``Learning deconvolution network for semantic segmentation,''
\newblock in {\em Proceedings of the IEEE ICCV}, 2015, pp. 1520--1528.

\bibitem{yu2015multi}
Fisher Yu and Vladlen Koltun,
\newblock ``Multi-scale context aggregation by dilated convolutions,''
\newblock {\em arXiv preprint arXiv:1511.07122}, 2015.

\bibitem{wu2018pocketflow}
Jiaxiang Wu, Yao Zhang, Haoli Bai, Huasong Zhong, Jinlong Hou, Wei Liu, and
  Junzhou Huang,
\newblock ``Pocketflow: An automated framework for compressing and accelerating
  deep neural networks,''
\newblock in {\em NIPS, Workshop}. 2018.

\end{thebibliography}
}

\end{document}